\documentclass[doublecol,figures]{epl2} % for 2 columns style with line numbers
\title{A stronger null hypothesis for crossing dependencies}
\author{R. Ferrer-i-Cancho\inst{1}\thanks{E-mail: \email{rferrericancho@cs.upc.edu}}}
\shortauthor{R. Ferrer-i-Cancho}

\institute{                    
  \inst{1} Complexity \& Quantitative Linguistics Lab \\
LARCA Research Group, \\
Departament de Ci\`encies de la Computaci\'o, \\
Universitat Polit\`ecnica de Catalunya, \\
Campus Nord, Edifici Omega Jordi Girona Salgado 1-3. \\
08034 Barcelona, Catalonia (Spain)
}
\abstract{
The syntactic structure of a sentence can be modeled as a tree where vertices are words and edges indicate syntactic dependencies between words. It is well-known that those edges normally do not cross when drawn over the sentence. Here a new null hypothesis for the number of edge crossings of a sentence is presented. That null hypothesis takes into account the length of the pair of edges that may cross and predicts the relative number of crossings in random trees with a small error, suggesting that a ban of crossings or a principle of minimization of crossings are not needed in general to explain the origins of non-crossing dependencies. Our work paves the way for more powerful null hypotheses to investigate the origins of non-crossing dependencies in nature. 
}

\begin{document}

\maketitle

\section{Introduction}

\label{introduction_section}

\begin{figure}
\begin{center}
\includegraphics[scale = 0.7]{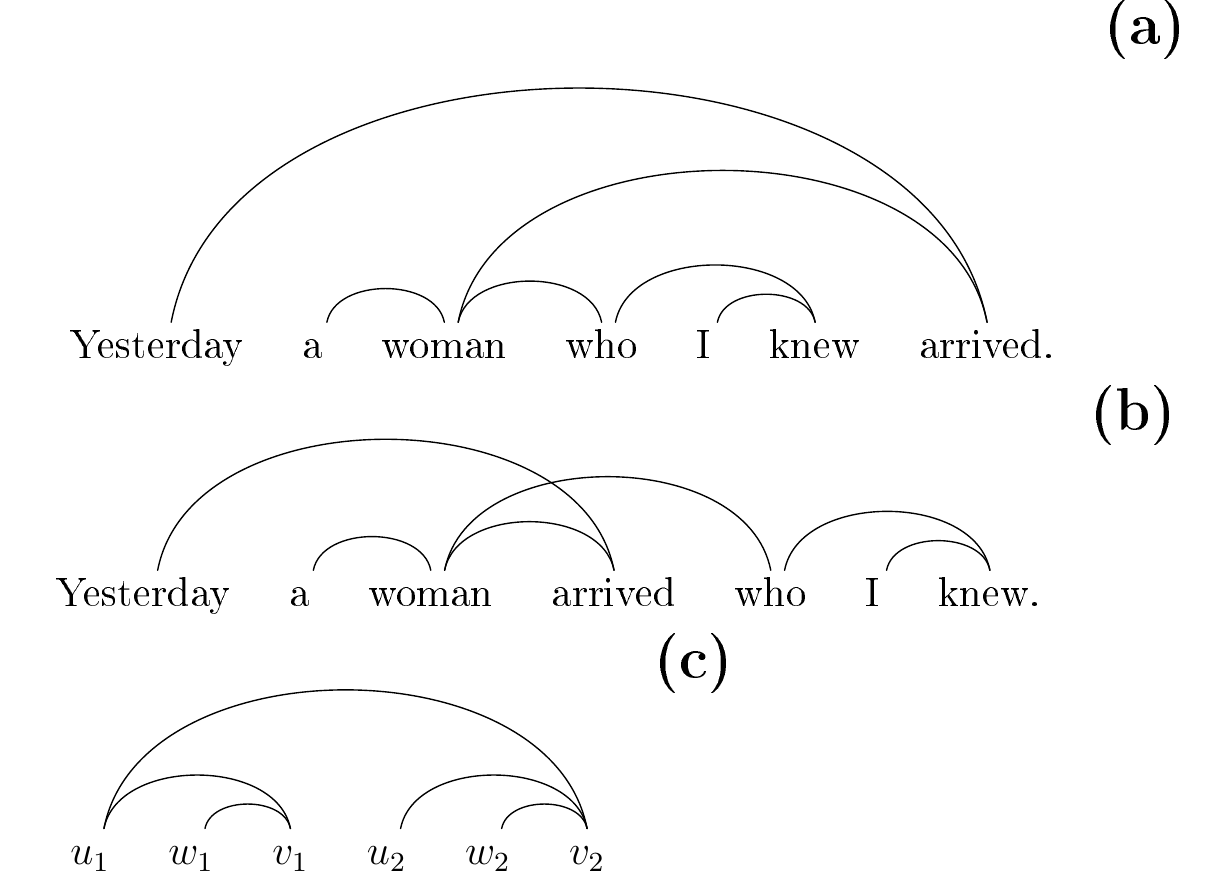}
\end{center}
\caption{\label{real_sentences_figure} 
(a) A sentence without crossings. (b) An alternative ordering yielding one crossing: the link $yesterday \sim arrived$ crosses the link $woman \sim who$ and vice versa. (c) An abstract structure. (a) and (b) are adapted from \cite{Levy2012a}. 
}
\end{figure}

The syntactic structure of a sentence can be defined as a network where vertices are words and edges indicate syntactic dependencies \cite{Melcuk1988,Hudson2007a} as in Fig.~\ref{real_sentences_figure}. The most common assumption is that this structure is a tree (an acyclic connected graph) ({\em e.g.}, \cite{Melcuk1988,Levy2012a}). % citation of McDonald2005a removed 
In the 1960s, a striking pattern of syntactic dependency trees of sentences was reported: dependencies between words normally do not cross when drawn over the sentence \cite{Lecerf1960a,Hays1964} ({\em e.g.}, Fig.~\ref{real_sentences_figure}). $C$, the number of different pairs of edges that cross, is small in real sentences. 
In Fig.~\ref{real_sentences_figure}, $C=0$ for sentence (a) and $C=1$ for sentence (b). Interestingly, the tree structure of both sentences is the same but $C$ varies, showing that $C$ depends on the linear arrangement of the vertices.

Imagine that $\pi(v)$ is defined as the position of the vertex $v$ in a linear arrangement of $n$ vertices (the 1st vertex has position 1, the second vertex has position 2 and so on...) and thus $1 \leq \pi(v) \leq n$. $u \sim v$ is used to refer to an edge formed by the vertices $u$ and $v$. The length of the edge $u \sim v$ in words is $d(u\sim v) = |\pi(u)-\pi(v)|$ (here $|...|$ is the absolute value operator). 
$s(u \sim v)$ and $e(u \sim v)$ are defined, respectively, as the initial and the end position of the edge $u \sim v$, {\em i.e.} $s(u \sim v)=min(\pi(u),\pi(v))$ and $e(u \sim v)=max(\pi(u),\pi(v))$.
$u_1 \sim v_1$ and $u_2 \sim v_2$ cross if and only if one of the following conditions is met 
\begin{itemize}
\item
$s(u_1 \sim v_1) < s(u_2 \sim v_2)$ and $s(u_2 \sim v_2) < e(u_1 \sim v_1)$ and $e(u_1 \sim v_1) < e(u_2 \sim v_2)$
\item
$s(u_1 \sim v_1) > s(u_2 \sim v_2)$ and $s(u_1 \sim v_1) < e(u_2 \sim v_2)$ and $e(u_2 \sim v_2) < e(u_1 \sim v_1)$.
\end{itemize}
It has been hypothesized that $C \approx 0$ in real sentences \cite{Melcuk1988,Liu2010a} could be due to a principle of minimization of the length of edges \cite{Ferrer2006d,Liu2008a,Morrill2009a,Ferrer2013b}. Although the minimization of 
\begin{equation}
D=\sum_{u\sim v} d(u\sim v)
\end{equation}
reduces crossings to practically zero \cite{Ferrer2006d}, this does not provide a full explanation about the low frequency of crossings in real sentences: (a) minimum $D$ does not imply $C=0$ \cite{Hochberg2003a}, (b) the actual value of $D$ in real sentences is located between the minimum and that of a random ordering of vertices \cite{Ferrer2004b} and (c) the word order that minimizes $D$ might be in a serious conflict with other linguistic or cognitive constraints \cite{Ferrer2014a}. Here the problem of the reduction of $D$ that is required  for explaining $C \approx 0$ in real sentences is avoided by means of a null hypothesis that predicts $C$ by considering the actual length of the edges that may cross. With this null hypothesis, one can shed light on a fundamental question: how much surprising it is that $C \approx 0$ given the lengths of edges? That null hypothesis is vital for the development of a general but minimal theory of crossing dependencies in nature. First, $C \approx 0$ in sentences might also be due to a ban of crossings by grammar \cite{Hudson2007a} % cite also Hudson1984
or a principle of minimization of $C$ \cite{Liu2008a}.
Second, crossings have also been investigated in networks of nucleotides \cite{Chen2009a}.
% where vertices are occurrences of nucleotides $A$, $G$, $U$, and $C$ and edges are Watson-Crick ($A$-$U$, $G$-$C$) and $U$-$G$ base pairs \cite{Chen2009a}.
Here it will be shown that a simple null hypothesis based on actual dependency lengths would suffice {\em a priori} for predicting $C \approx 0$ in short enough sentences.

\section{Crossing theory}

\subsection{The expected number of crossings}

$C(u \sim v)$ is defined as the number of edge crossings where the edge formed by $u$ and $v$ is involved. $C$ can be defined as 
\begin{equation}
C = \frac{1}{2} \sum_{u\sim v} C(u \sim v),
\end{equation}
where the $1/2$ factor is due to the fact that if two edges $u_1 \sim v_1$ and $u_2 \sim v_2$ cross, their crossing will be counted twice, one through $C(u_1 \sim v_1)$ and another through $C(u_2 \sim v_2)$.
$C(u_1 \sim v_1)$ can be defined as 
\begin{equation}
C(u_1 \sim v_1) = \sum_{u_2 \sim v_2, \{ u_1, v_1 \} \cap \{ u_2, v_2 \} = \emptyset} C(u_1 \sim v_1, u_2 \sim v_2),
% u_2 \neq u_1,v_1, ~ v_2 \neq u_1,v_1
\label{crossings_of_an_edge_equation}
\end{equation}
where $C(u_1 \sim v_1, u_2 \sim v_2)$ indicates if $u_1,v_1$ and $u_2,v_2$ define a couple of edges that cross, {\em i.e.} $C(u_1 \sim v_1, u_2 \sim v_2) = 1$ if they cross, $C(u_1 \sim v_1, u_2 \sim v_2) = 0$ otherwise.
Applying the definition of $C(u \sim v)$ in eq.~(\ref{crossings_of_an_edge_equation}), $C$ becomes
\begin{equation}
C = \frac{1}{2} \sum_{u_1 \sim v_1} \sum_{u_2 \sim v_2, \{ u_1, v_1 \} \cap \{ u_2, v_2 \} = \emptyset} C(u_1 \sim v_1, u_2 \sim v_2). 
\end{equation}
Suppose that the vertices are arranged linearly at random (being all the permutations of the vertex sequence equally likely). Then, the expectation of $C$ is
\begin{widetext}
\begin{equation}
E[C] = \frac{1}{2} \sum_{u_1 \sim v_1} \sum_{u_2 \sim v_2, \{ u_1, v_1 \} \cap \{ u_2, v_2 \} = \emptyset} E[C(u_1 \sim v_1, u_2 \sim v_2)].
\label{expected_crossings_equation}
\end{equation}
\end{widetext}
\begin{floatequation}
\mbox{\textit{see eq.~(\ref{expected_crossings_equation})}}
\end{floatequation}
As $C(u_1 \sim v_1, u_2 \sim v_2)$ is and indicator variable, $E[C(u_1 \sim v_1, u_2 \sim v_2)]$ can be replaced by $p(cross)=1/3$, the probability that two arbitrary edges that to not share any vertex cross when their vertices are arranged linearly at random, which yields \cite{Ferrer2013d} 
\begin{equation}
E_0[C] = C_{max}/3
\label{expected_crossings0_equation}
\end{equation}
with 
\begin{equation}
C_{max} = \frac{n}{2}\left(n - 1 - \left< k^2 \right>\right)
\label{potential_crossings_equation}
\end{equation}
being the number of edge pairs that can potentially cross and $\left< k^2 \right>$ the degree 2nd moment of the tree \cite{Ferrer2013b}.
$\left< k^2 \right>$ is the mean of squared degrees, i.e. 
\begin{equation}
\left< k^2 \right> = \sum_v k_v^2, 
\end{equation} 
where $k_v$ is the degree of vertex $v$. In uniformly random labeled trees, the expected $\left< k^2 \right>$ is \cite{Noy1998a,Moon1970a}  
\begin{equation}
E\left[\left<k^2 \right>\right] = \left(1-\frac{1}{n}\right)\left(5 - \frac{6}{n}\right).
\label{degree_2nd_moment_null_hypothesis_equation}
\end{equation}
Thus, the expectation of $E_0[C]$ for those trees is 
\begin{eqnarray}
E[E_0[C]] & = & \frac{n}{6} \left( n-1-E \left[ \left< k^2 \right> \right] \right) \nonumber \\ 
          & = & \frac{n^2}{6} - n + \frac{11}{6} - \frac{1}{n}. \label{expected_crossings_in_random_labelled_trees_equation}
\end{eqnarray}
This analytical result is easy to check numerically by generating random linear arrangements of vertices of random trees with the procedure in Fig.~\ref{procedure_figure}.

\begin{figure}
\fbox{
\parbox{8cm}{
\small
\begin{itemize}
\item
Assume that the vertices are labeled with integers from $1$ to $n$. 
\item
Produce a uniformly random spanning tree with the Aldous-Brother algorithm \cite{Aldous1990a,Broder1989a}, assuming a complete graph as the basis of the random walk.
\item
Take vertex labels as vertex positions ($\pi(v) = v$ for every vertex $v$).  
\end{itemize}
}
}
\caption{\label{procedure_figure} Procedure to generate a random labeled tree and a random linear arrangement of its vertices. }
\end{figure}

\begin{widetext}
\begin{eqnarray}
E[C|d] & = &
\frac{1}{2} \sum_{u_1 \sim v_1} \sum_{u_2 \sim v_2, \{ u_1, v_1 \} \cap \{ u_2, v_2 \} = \emptyset} E[C(u_1 \sim v_1, u_2 \sim v_2) | d] \\
             & = &  
\frac{1}{2} \sum_{u_1 \sim v_1} \sum_{u_2 \sim v_2, \{ u_1, v_1 \} \cap \{ u_2, v_2 \} = \emptyset} p(u_1 \sim v_1 \mbox{~and~} u_2 \sim v_2 \mbox{~cross}|d).
\label{expected_crossings_full_equation}
\end{eqnarray}
\end{widetext}
 
Here we aim to improve $E_0[C]$ introducing information about the actual length of the dependencies.
Suppose that 
\begin{equation}
p(u_1 \sim v_1 \mbox{~and~} u_2 \sim v_2 \mbox{~}cross|d)
\end{equation}
is the probability that the edges $u_1 \sim v_1$ \mbox{~and~} $u_2 \sim v_2$ cross in a random linear arrangement of vertices where edge lengths are given by the function $d$ above.
Then, $E[C|d]$, the expected number of crossings given full knowledge about edge lengths, can be defined as
\begin{floatequation}
\mbox{\textit{see eq.~(\ref{expected_crossings_full_equation})}}
\end{floatequation}
The calculation of $E[C|d]$ for a given sentence is not straightforward: it requires the calculation of all the permutations of the words of the sentence preserving the edge lengths of the original sentence. Besides, $E[C|d]$ makes a prediction about the crossings of a dependency tree involving a lot of information: the edges of the tree and their length. In contrast, $E_0[C]$ can be computed just from knowledge about the degree sequence or simply the values of $n$ and $\left<k^2 \right>$, as eqs.~(\ref{expected_crossings0_equation}) and (\ref{potential_crossings_equation}) indicate. Here we aim to predict the number of crossings reducing the computational and informational demands of $E[C|d]$ while beating the predictions of $E_0[C]$. 

$p(cross|d(u_1\sim v_1),d(u_2\sim v_2))$ is defined as the probability that two edges that are arranged linearly at random cross knowing that their lengths are $d(u_1\sim v_1)$ and $d(u_2\sim v_2)$ and that they do not share any vertex.
Replacing 
\begin{equation}
p(u_1 \sim v_1 \mbox{~and~} u_2 \sim v_2 \mbox{~}cross|d)
\end{equation}
by $p(cross|d(u_1\sim v_1),d(u_2\sim v_2))$ in eq.~{\ref{expected_crossings_full_equation}}, one obtains
\begin{widetext}
\begin{equation}
E_2[C] =  
\frac{1}{2} \sum_{u_1 \sim v_1} \sum_{u_2 \sim v_2, \{ u_1, v_1 \} \cap \{ u_2, v_2 \} = \emptyset} p(cross|d(u_1 \sim v_1),d(u_2 \sim v_2)).
\label{expected_crossings2_equation}
\end{equation}
\end{widetext}
\begin{floatequation}
\mbox{\textit{see eq.~(\ref{expected_crossings2_equation})}}
\end{floatequation}
$E_x[C]$ refers to an approximation to the expected value of $C$ knowing the length of $x$ edges in every potential crossing (giving priority to the knowledge about the lengths of the pair of edges that may cross in every potential crossing as in eq.~(\ref{expected_crossings2_equation})). $E_2[C]$ is an approximation to $E[C|d]$ that is based on a stronger null hypothesis than that of $E_0[C]$ for the probability that two edges cross. $E_0[C]$ and $E_{n-1}[C]$ are true expectations (notice $E_{n-1}[C] = E[C|d]$). While $E[C|d]$ conditions globally with the function $d$, i.e. the same conditioning for every pair of edges that may cross, $E_2[C]$ conditions locally with two edge lengths that depend on the pair of edges under consideration (Eq. \ref{expected_crossings_full_equation} versus Eq. \ref{expected_crossings2_equation}).
In the remainder of the article two virtues of $E_2[C]$ over $E[C|d]$ will be shown. First, $E_2[C]$ is easier to calculate. Second, it predicts $C$ with small error in spite of discarding, for every pair of edges that may potentially cross, the lengths of other edges. The point is: if such a rough but simple predictor of crossing works, is it necessary to believe that crossings are forbidden by grammars \cite{Hudson2007a}
or postulate an independent principle of minimization of $C$ \cite{Liu2008a}?   

\subsection{The probability that two edges cross knowing their lengths}

The set $S(n, d)$ is defined as the set of possible initial positions for an edge of length $d$ in a sequence of length $n$, {\em i.e.} 
\begin{equation}
S(n, d) = \{s| 1 \leq s \leq n - d\}.
\end{equation}
We say that $s_1$ and $s_2$ are a valid pair of initial positions if they define the initial positions of two edges 
that have lengths $d_1$ and $d_2$, respectively, and that do not share vertices, {\em i.e.} $s_1 \in S(n, d_1)$, $s_2 \in S(n, d_2)$ and $\{ s_1, s_1+d_1 \} \cap \{ s_2, s_2+d_2 \} = \emptyset$. 

\begin{figure}
\begin{center}
\includegraphics[scale = 0.75]{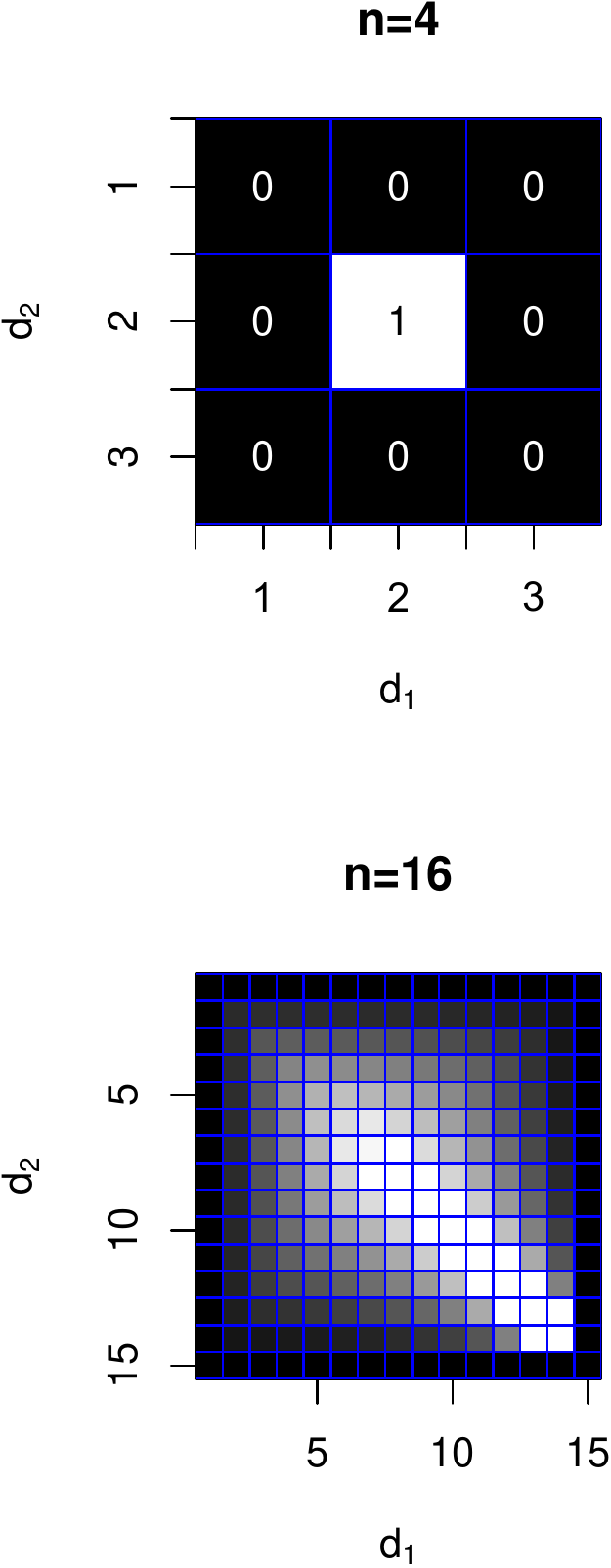}
% produced with:
% - plot_probability_given_two_distances_2_panels_vertical.R
% - the command pdfcrop probability_given_two_lengths.pdf
% - errors in axis labels corrected with Acrobat
\end{center}
\caption{\label{probability_of_crossing_given_two_lengths_figure} $p(cross | d_1, d_2)$, the probability that two edges cross when arranged linearly at random knowing their lengths ($d_1$ and $d_2$) and that they do not share vertices. Brightness is proportional to $p(cross | d_1, d_2)$ (black for $p(cross | d_1, d_2)=0$ and white for $p(cross | d_1, d_2) = 1$). 
$n$ is the number of vertices ($C>0$ needs $n \geq 4$ \cite{Ferrer2013b}). }
\end{figure}

$p(cross = 1|d_1,d_2)$ can be defined as a proportion, {\em i.e.} 
\begin{equation}
p(cross | d_1, d_2) = \frac{|\alpha(d_1,d_2)|}{|\beta(d_1,d_2)|},
\label{probability_of_crossing_given_two_lengths_equation}
\end{equation}
where here $|..|$ is the cardinality operator, $\alpha(d_1,d_2)$ is the set of valid pairs of initial position of two edges of lengths $d_1$ and $d_2$ that involve a crossing and $\beta(d_1,d_2)$ is simply the set of valid pairs of initial positions of edges of lengths $d_1$ and $d_2$. More formally, 
\begin{eqnarray}
\beta(d_1, d_2) = \{s_1,s_2| s_1 \mbox{~and~} s_2 \mbox{~are valid initial positions} \}
\end{eqnarray}
and 
\begin{widetext}
\begin{eqnarray}
\alpha(d_1, d_2) & = & \{s_1,s_2| s_1 \mbox{~and~} s_2 \mbox{~are valid initial positions and} \nonumber \\
                 &   & (s_1 < s_2 \mbox{~and~} s_2 < s_1 + d_1 \mbox{~and~} s_1 + d_1 < s_2 + d_2) \mbox{~or~} \nonumber \\
                 &   & (s_1 > s_2 \mbox{~and~} s_1 < s_2 + d_2 \mbox{~and~} s_2 + d_2 < s_1 + d_1) \}. \label{alpha_equation}
\end{eqnarray}
\end{widetext}
\begin{floatequation}
\mbox{\textit{see eq.~(\ref{alpha_equation})}}
\end{floatequation}
The definition of $\alpha(d_1, d_2)$ is based on an adapted version of the formal definition of crossing in the introduction section
% ~\ref{introduction_section} 
(notice that $e(u\sim v)= s(u\sim v) + d(u\sim v)$).
Fig.~\ref{probability_of_crossing_given_two_lengths_figure} shows $p(cross | d_1, d_2)$ for two different number of vertices. 
If $\beta(d_1, d_2) = 0$ then $\alpha(d_1, d_2)=0$ and then $p(cross | d_1, d_2)$ is undefined (notice that $\beta(n-1,n-1)=\beta(n-2,n-1)=\beta(n-1,n-2)=0$). If that happens, the reasonable convention that $p(cross | d_1,d_2) = 0$ is adopted. 
The order of edge length information is irrelevant, {\em i.e.} $p(cross | d_1, d_2) = p(cross | d_2, d_1)$ as Fig.~\ref{probability_of_crossing_given_two_lengths_figure} shows. Some crossings are impossible {\em  a priori}, {\em i.e.} $p(cross | 1, d_2) = p(cross | n-1, d_2) = 0$ and some others are unavoidable, {\em e.g.}, $p(cross | n-2,n-2) = 1$ (we are assuming $n\geq 4$).

$p(cross)$ and $p(cross | d_1, d_2)$ are related through
\begin{equation}
\sum_{d_1=1}^{n-1} \sum_{d_2=1}^{n-1} p(cross | d_1, d_2)p(d_1,d_2) = p(cross),
\label{conditional_versus_unconditional_probability_of_crossing_equation}
\end{equation}
where $p(d_1,d_2)$ is the probability that a random linear arrangement of four different vertices, {\em i.e.} $u_1$,$v_1$,$u_2$ and $v_2$, produces $|\pi(u_1)-\pi(v_1)| = d_1$ and $|\pi(u_2)-\pi(v_2)| = d_2$.

\section{Results}

\begin{largetable}
\caption{\label{predictions_table} The properties and predictions of crossings for the sentences in Fig.~\ref{real_sentences_figure}. $n$ is the number of vertices (sentence length in words), $\left< k^2 \right>$ is the degree 2nd moment, $C_{max}$ is the potential number of crossings, $C_{true}$ and $\bar{C}_{true}$ are, respectively, the absolute and the relative actual number of crossings. $E_0[...]$ is the expectation of crossings ignoring edge lengths and $E_2[...]$ is an approximation to the expectation knowing the lengths of edges. Numbers were rounded to leave two significant decimals. % (zeroes are not shown).
} 
\begin{center}
\item[]\begin{tabular}{@{}llllllllll}
Example & $n$ & $\left< k^2 \right>$ & $C_{max}$ & $C_{true}$ & $E_0[C]$ & $E_2[C]$ & $\bar{C}_{true}$ & $E_0[\bar{C}]$ & $E_2[\bar{C}]$\\ 
\hline
Fig. 1 (a) & 7 & 3.4 & 9 & 0 & 3 & 0.57 & 0 & 0.33 & 0.063 \\
Fig. 1 (b) & 7 & 3.4 & 9 & 1 & 3 & 1.5 & 0.11 & 0.33 & 0.17 \\
\end{tabular}
\end{center}
\end{largetable}

The relative number of crossings is defined as $\bar{C}_{true} = C_{true}/C_{max}$ and thus $E_x[\bar{C}]= E_x[C]/C_{max}$.
Table~\ref{predictions_table} shows that $E_2[...]$ makes better predictions about the (absolute or relative) number of crossings than $E_0[...]$ for the real syntactic dependency trees in Fig.~\ref{real_sentences_figure}. $\bar{C}_{true}$ and $E_x[\bar{C}]$ allow for a fairer comparison of the real number of crossings and its predictions as they measure crossings in units of the potential number of crossings. 
We wish to investigate if $E_x[\bar{C}]$ might shed light on the small number of crossings of real sentences abstracting away from the details of a concrete language, in the spirit of a long tradition of research on crossing dependencies \cite{deVries2012a,Christiansen1999a}. 
% Against English linguistics, Evans & Levinson article in BBS
Our language neutral perspective is not based on the analysis of real syntactic dependency trees but those of uniformly random labeled trees whose vertex labels are distinctive numbers from $1$ to $n$ that also represent the positions of the vertices, {\em i.e.} $\pi(v) = v$. Here we aim to compare the capacity of $E_0[\bar{C}]$ and $E_2[\bar{C}]$ to predict $\bar{C}_{true}$, the real number of a crossings in uniformly random labeled trees, when $C_{true}$ is small ($C_{true} \leq 3$) as in real sentences \cite{Lecerf1960a,Hays1964}. The relative error of the prediction is defined as 
\begin{eqnarray}
\Delta_x & = & E_x[\bar{C}] - \bar{C}_{true} \nonumber \\
         & = & (E_x[C] - C_{true})/C_{max}.
\end{eqnarray}
For every sentence of length $n \geq 4$ (because $C>0$ needs it \cite{Ferrer2013b}), an ensemble of $R=10^4$ uniformly random labeled trees with $C_{true} \leq 3$ was generated (a) following the procedure in Fig.~\ref{procedure_figure} and (b) rejecting random trees yielding $C_{true}>3$ till  the desired size $R$ was reached. For every relevant value of $C_{true}$ ($0 \leq C_{true} \leq 3$), the mean $\Delta_2$ was calculated over all configurations where $C_{max} > 0$ ($C_{max} = 0$ is only achieved by star trees \cite{Ferrer2013b}). $n_{max}=20$ was the maximum sentence length considered due to the explosion of rejections as $n$ increases. The space of possible trees is huge (there are $n^{n-2}$ labeled trees of $n$ vertices \cite{Cayley1889a}) and trees with $C_{true} \leq 3$ have a number of crossings that is unexpectedly low for that class of random trees (recall eq.~(\ref{expected_crossings_in_random_labelled_trees_equation})).     
These considerations notwithstanding, $n_{max}$ covers the average length of English sentences (about 17.8 words \cite[pp. 37-55]{Leech2007a}),  
% pages 37-55
% and that of many other languages (cf. Table 1 of \cite{Buchholz2006a}). % compte: inclouen signes de puntuacio
and that of other languages \cite{Ferrer2004b}.

\begin{figure*}[t]
\begin{center}
\includegraphics[scale = 0.8]{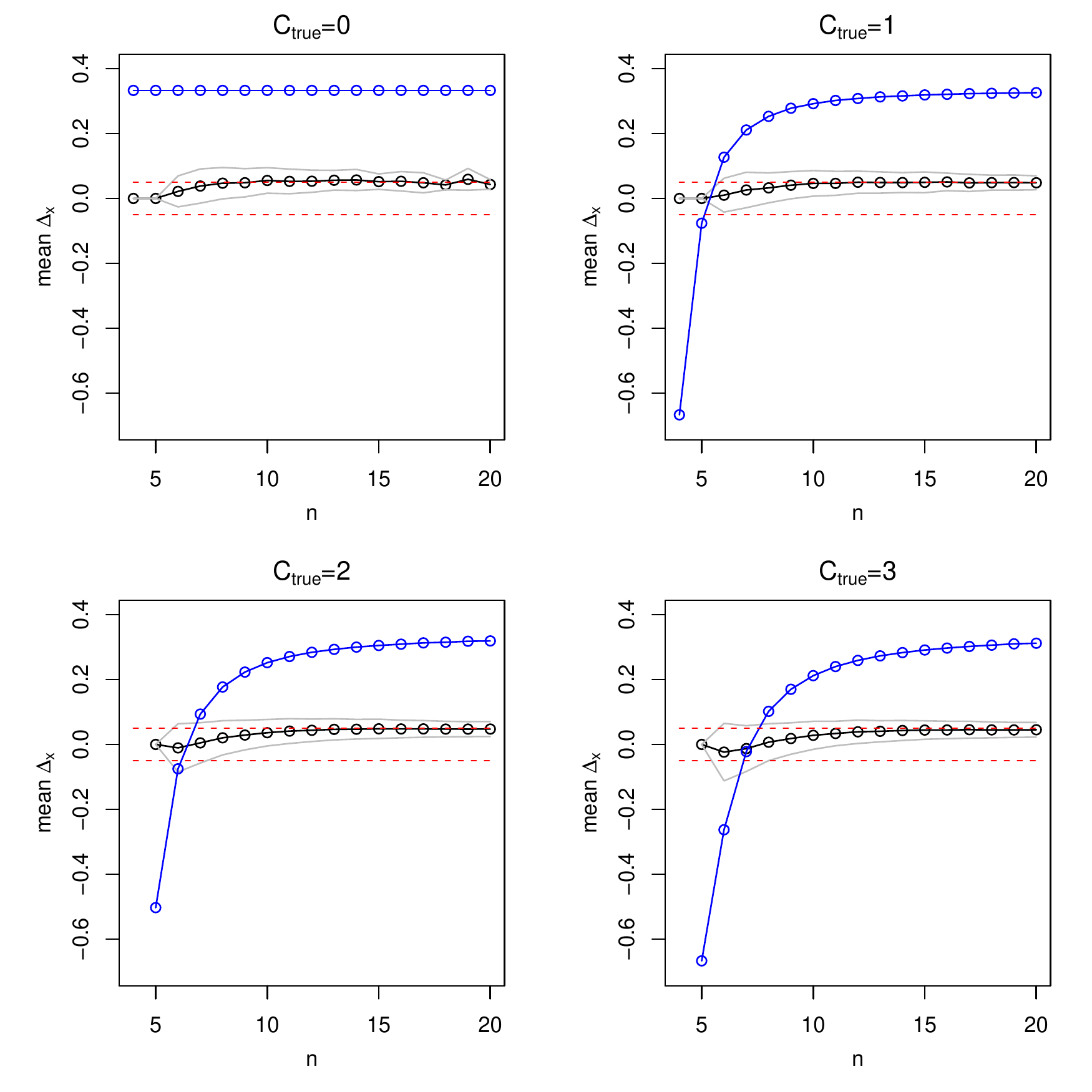}
\end{center}
\caption{\label{predicted_crossings_error_figure} The average relative error $\Delta_x$ as function of the number of vertices $n$ of the random trees conditioning on $C_{true}$ (black for $x=2$ and blue for $x=0$). The mean $\Delta_2$ is surrounded by two boundary gray lines: one standard deviation above and one standard deviation below. The two red dashed lines are a guide to the eye for $\Delta_x = \pm 0.05$. $C_{true}>1$ is impossible for $n < 4$ \cite{Ferrer2013b}. }
\end{figure*}

Fig.~\ref{predicted_crossings_error_figure} shows the mean $\Delta_x$ over ensembles of random trees with $C_{true} \leq 3$ indicating both $E_0[\bar{C}]$ and $E_2[\bar{C}]$ overestimate  
$\bar{C}_{true}$ in general. While $\Delta_2$ is small, {\em i.e.} of the order of $5\%$, 
$\Delta_0$ converges to $1/3$ as expected from the fact that 
\begin{eqnarray}
\Delta_0 & = & (C_{max}/3 - C_{true})/C_{max} \nonumber \\ 
         & = & 1/3 - C_{true}/C_{max},
\end{eqnarray}
which yields $\Delta_0 \approx 1/3$ for sufficiently large $n$ and $C_{true}$ small. 

\section{Discussion}

It has been shown that $E_2[\bar{C}]$ is able to predict the actual relative number of crossings in random unlabeled trees. This is not very surprising: edge length does give information on how likely edges are to cross. What is not straightforward is that a method that estimates crossings based exclusively on local dependency length information (just on the length of the pair of edges that can potentially cross) is able to make predictions with a small relative error in trees of the size of real sentences. Our finding has important consequences for language research: it suggests that there is no need {\em a priori} for banning crossings by grammar \cite{Hudson2007a} % cite also Hudson1984
or minimizing $C$ \cite{Liu2008a} to explain $C \approx 0$ in short enough sentences. This is consistent with the view that syntactic constraints, in general, do not imply an internally represented grammar \cite{Christiansen1999a}.

However, the predictive power of $E_2[\bar{C}]$ decreases slightly as the number of vertices increases (Fig.~\ref{predicted_crossings_error_figure}). The reason is very simple: $E_2[...]$ departs from an estimation of the probability that two edges cross that is based exclusively on their lengths, thus discarding the length of other edges. $p(cross | d_1, d_2)$ neglects the length of $n-3$ edges. As $n$ increases, the amount of information discarded increases and predictions worsen. In the tree in Fig.~\ref{real_sentences_figure} (c), the only pairs of edges that could cross in the sense of $p(cross | d_1, d_2)>0$ (i.e. if dependency lengths of other edges were ignored) are $u_1 \sim v_1$ and $u_2 \sim v_2$ (recall that edges of length 1 or $n-1$ cannot produce crossings). Eq.~(\ref{probability_of_crossing_given_two_lengths_equation}) gives $p(cross | d_1 = d_2 = 2) = 0.75$ but $p(C(u_1 \sim v_1, u_2 \sim v_2) = 1| d(u_1\sim v_1) = d(u_2\sim v_2) = 2, d(u_1 \sim v_2)=5) = 0$ ($d(u_1 \sim v_2)=5$ can only be achieved placing $u_1$ and $v_2$ at the ends of the sequence, which turns $C(u_1 \sim v_1, u_2 \sim v_2) = 1$ impossible). For this reason, $E_{n-1}[\bar{C}]$, the expected relative number of crossings knowing all edge lengths in every potential crossing, should be investigated in the future. 
% In the meantime, $E_2[\bar{C}]$ offers a computationally simpler prediction that outperforms that of $E_0[\bar{C}]$.

% \begin{figure}
% \begin{center}
% \includegraphics[scale = 0.8]{artificial_dependency_tree}
% \end{center}
% \caption{\label{artificial_sentence_figure} An artificial dependency tree.}
% \end{figure}

\acknowledgments
We are grateful to D. Blasi, R. Czech, E. Gibson and G. Morrill for helpful discussions.  This work was supported by the grant BASMATI
(TIN2011-27479-C04-03) from the Spanish Ministry of Science and Innovation.

\bibliographystyle{eplbib}

% \bibliography{../biblio/complex,../biblio/rferrericancho,../biblio/ling,../biblio/cs,../biblio/cl,../biblio/maths}

\end{document}